\documentclass[aps,pre,twocolumn,groupedaddress,floatfix,superscriptaddress]{revtex4}
\usepackage{dcolumn}
\usepackage{amsmath}
\usepackage{amssymb}
\usepackage{lipsum}
\usepackage{graphicx}
\usepackage{bm}
\usepackage[T1]{fontenc}
\usepackage{color}
\usepackage{url}
\usepackage[bf]{subfigure}
\usepackage{rotating}
\usepackage{soul}
\usepackage{scalefnt}
\usepackage{hyperref}
\usepackage{scalefnt}

\usepackage{perpage} 
\MakePerPage{footnote}

\usepackage{color}
\definecolor{redcolor}{rgb}{1.0,0.,0.}

\begin{document}

\title{Representation of texts as complex networks: a mesoscopic approach}

\author{Henrique Ferraz de Arruda}
\affiliation{Institute of Mathematics and Computer Science, University of S\~ao Paulo, S\~ao Carlos, SP, Brazil.}
\author{Filipi Nascimento Silva}
\affiliation{S\~ao Carlos Institute of Physics, University of S\~ao Paulo, S\~ao Carlos, SP, Brazil}
\author{Vanessa Queiroz Marinho}
\author{Diego Raphael Amancio$^*$}
\affiliation{Institute of Mathematics and Computer Science, University of S\~ao Paulo, S\~ao Carlos, SP, Brazil.}
\author{Luciano da Fontoura Costa}
\affiliation{S\~ao Carlos Institute of Physics, University of S\~ao Paulo, S\~ao Carlos, SP, Brazil}

\begin{abstract}
Statistical techniques that analyze texts, referred to as text analytics, have departed from the use of simple word count statistics towards a new paradigm. Text mining now hinges on a more sophisticated set of methods, including the representations in terms of complex networks. While well-established word-adjacency (co-occurrence) methods successfully grasp syntactical features of written texts, they are unable to represent important aspects of textual data, such as its topical structure, i.e. the sequence of subjects developing at a mesoscopic level along the text. Such aspects are often overlooked by current methodologies. In order to grasp the mesoscopic characteristics of semantical content in written texts, we devised a network model which is able to analyze documents in a multi-scale fashion. In the proposed model, a limited amount of adjacent paragraphs are represented as nodes, which are connected whenever they share a minimum semantical content.  To illustrate the capabilities of our model, we present, as a case example, a qualitative analysis of ``Alice's Adventures in Wonderland''. We show that the mesoscopic structure of a document, modeled as a network, reveals many semantic traits of texts.  Such an approach paves the way to a myriad of semantic-based applications.  In addition, our approach is illustrated in a machine learning context, in which texts are classified among real texts and randomized instances.
\end{abstract}

\maketitle

\setcounter{secnumdepth}{1}

\section{Introduction}
The availability of an ever growing amount of data brought up by the age of information has strongly impacted science, giving rise to a novel perspective on data analysis. The use and development of systematic approaches to analyze data has already become mandatory in a wide range of knowledge areas, such as physics~\citep{boccaletti2006complex}, biology~\citep{barabasi2004network,de2015framework}, medicine~\citep{barabasi2011network} and even humanities~\citep{moreno2004dynamics,kalimeri2015word}. This also includes techniques devoted to the systematic analysis of texts, known as \emph{text mining}~\citep{Manning:1999:FSN:311445}. Traditionally, approaches involving text analytics were solely based on simple statistics considering mostly the frequency of words~\citep{nahm2002text, altmann2009beyond}, which are, in general, suitable for the task of text classification~\citep{joachims2001statistical}. However, more sophisticated methods have been devised for complex tasks, such as to quantify the words relevance~\citep{ramos2003using,hotho2005brief} in a document. These techniques can be employed to detect, for instance, important topics in a given text~\citep{blei2003latent, alsumait2008line}. Even more challenging are the methods used to study the relationships among words or topics in a document or a set of documents. This kind of analysis can be undertaken by considering semantic similarities~\citep{landauer1998introduction} or linguistic characteristics~\citep{hatzivassiloglou2000investigation}. By using these new techniques, many other applications could be achieved, e.g., automatic summarization~\citep{chang2009latent}, {event summary from many doccuments~\citep{wei2016overlapping}}, sentiment analysis~\citep{maas2011learning} or authorship detection~\citep{chen2011authorship}. Applications that illustrate the temporal dynamics~\citep{liu2013story,tanahashi2012design,prado2016temporal} are also important. In these works, texts or movies are analyzed according to the way entities (mainly characters) interact through time. Recently, 
\cite{reagan2016emotional} investigated how the emotional content evolves in a story. Moreover, text datasets can also be analyzed in terms of the relationships among their elements, such as words and paragraphs. So, texts can be regarded as a complex structure and, therefore, be suitably represented in terms of complex networks.

A well-known approach to construct complex networks from texts is the word-adjacency (or co-occurrence) technique~\citep{Amancio20124406,PhysRevE.91.032810}, which is based on connecting pairs of words that are immediately adjacent. The strategy of mapping texts according to co-occurrence relationships is a simplification of networks formed by syntactical links~\citep{PhysRevE.69.051915}. Despite this seeming limitation, word adjacency networks have been employed successfully to address a great variety of natural language processing problems. This includes sentiment analysis~\citep{Feldman:2013:TAS:2436256.2436274}, authorship detection~\citep{Mehri20122429,7140830,1742-5468-2015-3-P03005}, stylometry~\citep{10.1371/journal.pone.0136076}, text classification~\citep{de2016using}, word sense disambiguation~\citep{Mihalcea:2004:PSN:1220355.1220517,silva2012word,0295-5075-98-1-18002}, text summarization~\citep{Antiqueira2009584,Amancio20121855}, machine translation~\citep{PhysRevE.80.026103,Amancio20121855} and others.

Perhaps the most critical disadvantage associated with the word adjacency approach is its inability to portray the topical structure presented in many texts.   The topical structure of a text is expected to naturally emerge from its network representation through a pronounced heterogeneous macro-structure.  However, this hardly happens on typical co-occurrence networks, which present no community structure~\citep{topicseg}. This suggests that the co-occurrence representation does not effectively capture the information at the mesoscopic structure of the text, such as topics and subtopics. In addition, the information regarding the temporal evolution along a text is also overlooked in co-occurrence networks.

In order to address the above limitations, we propose a mesoscopic representation of texts, where a node represents a large context, e.g. a set of adjacent sentences or paragraphs. More specifically, in our approach each node corresponds to $\Delta$ subsequent paragraphs.  The relationship between these nodes is then established by a similarity criteria. As such, edges are created whenever a large number of words is shared between two nodes. Note that, by doing so, the network structure becomes more dependent on how the author approaches the topics along the text.  As we shall show, the proposed  representation is able to reflect the semantic complexity of texts, a feature that cannot be straightforwardly obtained in traditional word adjacency networks.

This manuscript is organized as follows: Section~II describes our approach to create the mesoscopic network from a given document. Section~III describes a case study of our approach. Section~IV illustrates the mesoscopic approach in a machine learning context. Finally, Section~V concludes our paper and suggests perspectives for further studies.

\section{Methods}\label{methods_section}
This section describes the procedure to obtain mesoscopic complex networks from texts, which include books and other documents with paragraph structure. Here, we also briefly present the technique employed to visualize these networks.

\subsection{From texts to networks}
In recent years, a new set of techniques has been introduced to create networks from documents, which takes into account their mesoscopic structure~\citep{topicseg}. In that work, the networks are generated by connecting words existing in the same context, which is defined in terms of a fixed window length. This approach was able to produce modular networks, with each community related to contextual topics or subtopics of the text~\citep{topicseg}. Even though the semantical organization of texts is captured by this representation, it is not straightforward to obtain the temporal evolution of the story being told.

Here, we extend the concepts introduced by \cite{topicseg} to derive a new technique to construct networks from texts. Our methodology addresses two important aspects typically overlooked by more traditional approaches: (a) the mesoscopic structure of a text and (b) its unfolding along timeunfolding along tim. To consider (a), instead of linking adjacent words, we use larger pieces of text as the basic representational unit. These pieces are connected according to the similarity among themselves. The temporal evolution of ideas and concepts is incorporated into our model because, by construction, successive nodes always result connected as a consequence of their shared content.

Henceforth, we consider an \emph{organized text} as a sequence of words delimitated by paragraphs. In our analysis, the paragraphs can be retained from the text, or can be inferred from the text own structure, for instance, by considering sequences with a fixed number of words.

Our approach starts with a pre-processing step typically employed for semantical-based text analysis. First, punctuation marks and numbers are removed. We also discard words conveying little contextual meaning, i.e. the \emph{stopwords}.  Examples of stopwords are articles and prepositions. If a lemmatization technique~\citep{Manning:1999:FSN:311445} is available for the language being considered, it is used to normalize concepts. In this step, words are reduced to their canonical forms, so that inflections in verbal tense, number, case or gender are disregarded. For example, the sentence ```{\it Oh, I've had such a curious dream!' said Alice}'' becomes  ``{\it curious dream say alice}'', after being pre-processed. Next, we employ the \emph{tf-idf} (term frequency-inverse document frequency) technique~\citep{Manning:1999:FSN:311445}, which defines a map $\text{tf-idf}(w,d,D)$ quantifying the importance of each word $w$ in a given document $d$ from a set of documents $D$. The $\text{tf-idf}(w,d,D)$ map is computed as
\begin{equation}
 \text{tf-idf}(w,d,D) = \text{tf}(w,d) \times \text{idf}(w,D),
\end{equation}
where $\text{tf}(w,d)$, the term-frequency component, accounts for the relevance of $w \in d$ and $\text{idf}(w,D)$, the inverse document frequency, quantifies the frequency of $w$ in all $d \in D$.  Many variations of both tf and idf terms have been proposed~\citep{Manning:1999:FSN:311445}. In this paper, we consider $\text{tf}(w,d)$ as the raw frequency of a given word $w$ in a document $d$. The $\text{idf}(w,D)$ is calculated as
\begin{equation} \label{tfeq}
 \textrm{idf}(w,D) = \log\Bigg{(}\frac{|D|}{f_w}\Bigg{)},
\end{equation}
where $|D|$ is the total number of documents in $D$ and $f_w$ is the number of documents in which $w$ occurs at least once.

\begin{figure*}
  \begin{center}
  \includegraphics[width=0.85\linewidth]{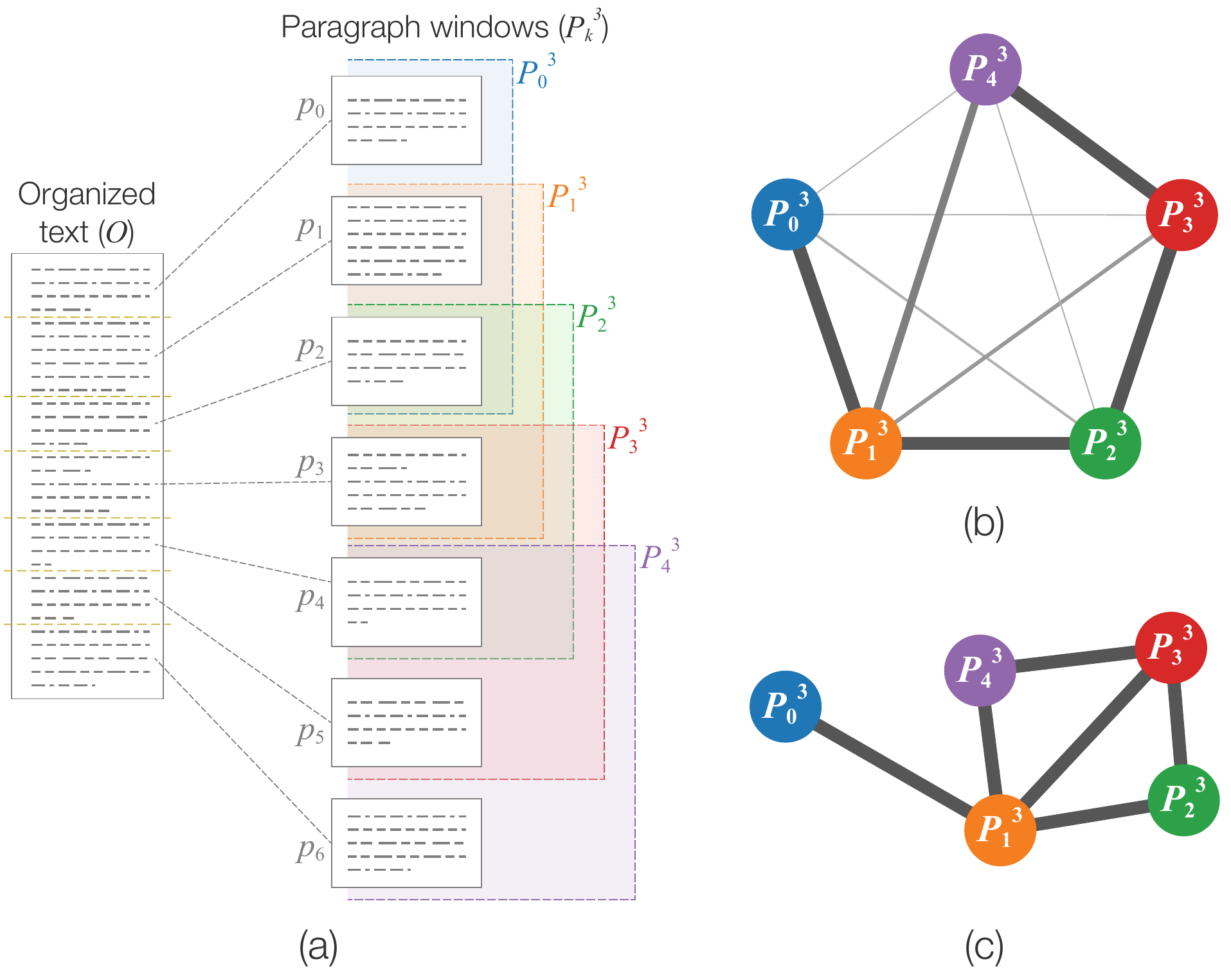}
  \caption{Illustration of the presented methodology. Initially, the text is organized in sets of subsequent and overlapping windows $P_{k}^{3}$, each containing $3$ structural paragraphs, as shown in (a). Next, the cosine similarity is calculated among all pairs of text windows (illustrated by the width of the lines in b). The mesoscopic network is obtained by maintaining only connections among pairs with similarity higher than a threshold value $T$. This is illustrated by the network visualization in (c).}
  ~\label{f:method}
  \end{center}
\end{figure*}

The mesoscopic network is generated from the preprocessed text, hereafter referred to as organized text $O$. The organized text $O$ consists of a sequence of paragraphs $O = (p_0,p_1,p_2\dots)$ with each paragraph $p_i$ comprising a sequence of words $p_i = (w_{i0},w_{i1},w_{i2}\dots)$. Differently from the co-occurrence model where nodes represent words, here, we map entire paragraphs or sequences of consecutive paragraphs as nodes. In particular, for a choice of window size $\Delta$, each possible subsequence comprising $\Delta$ paragraphs in $O$, $P_{k}^{\Delta}=(p_{k},p_{k+1},\dots p_{k+\Delta-1})$, is represented by a node in the devised mesoscopic network. Fig.~\ref{f:method}(a) illustrates the process of obtaining the nodes of the mesoscopic network.

The edges of the mesoscopic network are identified by calculating a contextual similarity measurement considering all pairs of sequences of paragraphs $P_{k}^\Delta$ in the investigated document. Here, we employed the traditional bag of words combined with the cosine similarity measurement~\citep{Manning:1999:FSN:311445}. Bearing in mind that the number of words in each paragraph can vary significantly, the cosine similarity was used because it does not depend on the length of the text chunks being compared~\citep{han2005dataMining}. First, for each considered sequence of paragraphs $P$, a vector $W_P$, spanning the same number of words present in $O$, is obtained from the $\text{tf-idf}(w, P, O)$ map applied to each word $w$ in $O$. Note that, when a certain word $w$ is not present in $P$, $\text{tf-idf}(w, P, O)=0$. The content similarity measurement $S(P_A,P_B)$ between two paragraph windows $P_A$ and $P_B$ is obtained using
\begin{widetext}
\begin{equation}
 S(P_A,P_B) = \frac{\sum\limits_{w\,\in\,O} {\text{tf-idf}(w, P_A, O) \times \text{tf-idf}(w, P_B, O)}}{\sqrt{\sum\limits_{w\,\in\,O}  {\text{tf-idf}(w, P_A, O)^2}} \sqrt{\sum\limits_{w\,\in\,O} {\text{tf-idf}(w, P_B, O)^2}}}.
\end{equation}
\end{widetext}
%
%
As a result, a fully connected network is created (see Fig.~\ref{f:method}(b)), in which the edge weights correspond to the similarity $S(P_A,P_B)$ among each pair of nodes. The final mesoscopic network is obtained by pruning the weakest connections, i.e. the links whose weight takes a value below a given  threshold $T$. After this procedure, edge weights are ignored, resulting in an unweighted network (see Fig.~\ref{f:method}(c)).

To better understand the overall structure of mesoscopic networks, we visualized the network structure using a technique based on force-directed nodes placement. In particular, we used a technique inspired on the Fruchterman-Reingold (FR)~\citep{Fruchterman1991Gr} algorithm, in which the network is regarded as a system of nodes behaving like particles that interact by the action of two types of forces: attractive forces, existing only between connected nodes, and repulsive forces, that exist between all pairs of nodes. By minimizing the energy of that system, the network organizes itself in a graphically appealing layout. This visualization technique naturally highlights many aspects of the topological structure of networks~\citep{Fruchterman1991Gr}.

\subsection{Results evaluation}
In order to show the potential of our networks to reflect the document story, we compared networks created from Real Texts (RT) with networks created from Shuffled Texts (ST), where clearly no story exists. The shuffled texts were created in a two-fold manner: obtained by shuffling words (SW) or paragraphs (SP) from real texts. To generate the SW version, all words from a given text were shuffled and the paragraphs were created with the same number of words as those in the original document. It is important to highlight that the number of paragraphs, their respective order and the number of words in each paragraph were preserved. In the second version of shuffled texts, SP, we shuffled all paragraphs from a given real text. Thus, the structure of each single paragraph is kept, but the new sequence of paragraphs may not generate a consistent, coherent story.

For each document, a single weighted mesoscopic network was created for each class (RT, SW, and SP). Consequently, the classes have the same number of networks. Considering the classes of text (RT, SW, and SP), for each weighted network, we generated unweighted networks from a set of thresholds ($T$). These thresholds were defined according to a given percentage of expected edges, so that edges with higher weights were maintained. After removing edges whose weights were below the threshold $T$, we used two measurements to compare the mesoscopic networks:
\begin{enumerate}
\item {\bf Clustering coefficient}:
this measurement is well known in complex networks analysis~\citep{watts1998collective} and it was used in many text classification applications~\citep{masucci2006network,sheng2009english,amancio2011comparing}. The clustering coefficient quantifies the fraction of loops of order three (i.e. triangles), for each network node and it is computed as
\begin{equation}
	C_i = \frac{N_\Delta(i)}{N_3(i)},
\end{equation}
where $N_\Delta(i)$ is the number of connected triangles in which node $i$ takes part and $N_3(i)$ is the number of connected triples, where $i$ is the central node;

\item {\bf Matching index}:  for each edge, this measure computes the similarity between the two nodes connected to the edge according to the number of common neighbors~\citep{Newman:2010:NI:1809753,sporns2003graph,kaiser2004edge}. In other words, this measurement quantifies the similarity between two network regions connected by an edge. This measurement is computed as
\begin{equation}
\mu_{i,j} = \frac{\sum_{k \neq i,j} a_{ik} a_{jk}}{\sum_{k \neq j} a_{ik} + \sum_{k \neq i} a_{jk}},
\end{equation}
where $a_{ij}$ is an element of the adjacency matrix, and  $a_{ij} = 1$ if nodes $i$ and $j$ are connected.
\end{enumerate}

The books were considered in their entirity.  As a consequence, the number of network nodes varies, which can influence many complex network measurements. As a solution for this problem, we analyzed the network in terms of local measurements of clustering and matching index.

In order to provide additional information about the text, the two measurements were calculated for all nodes/edges and sorted according to the text sequence, giving rise to a time series.  For the matching index, we created the time series by establishing the following order of edges:
\begin{equation} \nonumber
\{\mu_{0,0},\mu_{0,1},\ldots,\mu_{0,n-1},\mu_{1,0},\mu_{1,1}\ldots\mu_{1,n-1},\ldots,\mu_{n-1,n-1}\}.
\end{equation}
If there is no edge linking two nodes, the corresponding value in the  time series is not taken into account.

\section{Case study: Mesoscopic analysis of ``Alice's adventures in wonderland''}

In order to illustrate the potential of modeling real texts as mesoscopic networks, we applied our methodology to the well-kwown book ``Alice's Adventures in Wonderland''. This story revolves around the adventures of a little girl, called Alice, after she falls in a hole and arrives in an unknown fantasy world. The book was written in 1865 by Charles Lutwidge Dodgson under the pseudonym Lewis Carroll. It is divided into the following twelve chapters:
\begin{enumerate}
 \item Down the Rabbit-Hole
 \item The Pool of Tears
 \item A Caucus-Race and a Long Tale
 \item The Rabbit Sends in a Little Bill
 \item Advice from a Caterpillar
 \item Pig and Pepper
 \item A Mad Tea-Party
 \item The Queen's Croquet-Ground
 \item The Mock Turtle's Story
 \item The Lobster Quadrille
 \item Who Stole the Tarts?
 \item Alice's Evidence
\end{enumerate}

After the pre-processing steps had been undertaken, we chose a fixed window size  $\Delta = 20$ paragraphs. Two mesoscopic networks, $\mathcal{G}_1$ and $\mathcal{G}_2$, were constructed from the book with distinct thresholds to prune connections, $T_1=0.31$ and $T_2=0.18$, respectively. With these thresholds, only 5\% and 10\% of the edges, respectively, remained in the network.

We start the analysis of the mesoscopic structures by investigating the properties of the $\mathcal{G}_1$ network, which is simpler than $\mathcal{G}_2$. For this analysis, we consider a 2D visualization of $\mathcal{G}_1$, which is shown in Figure~\ref{f:aliceNetworks}. This visualization was obtained by employing the FR algorithm mentioned in Section~\ref{methods_section}. Because nodes sharing the same paragraphs become strongly connected among themselves, a pronounced chain-like structure naturally emerges on the mesoscopic network. In addition, this structure is related to the order of the nodes along the book. This property is better observed in Figure~\ref{f:aliceNetworks}(a), where the color of each node indicates its position along the text. In mesoscopic networks, connections among distant nodes indicate regions of high contextual similarity that are not a result of overlapping sequences of paragraphs. In these networks, the structure connects contextually similar regions of nodes which, by its turn, brings them closer along the chain-like structure of the network.

In order to better understand the relationship between the mesoscopic structure and the contextual information of the book, we segmented the obtained network according to the chapter organization of the book. This is visualized in Figure~\ref{f:aliceNetworks}(b), in which the chapter of each node is indicated by a color, according to the legend. Considering the connectivity among the chapters of the book, we derived the following observations:

\begin{figure*}[ht!]
  \begin{center}
  \includegraphics[width=\linewidth]{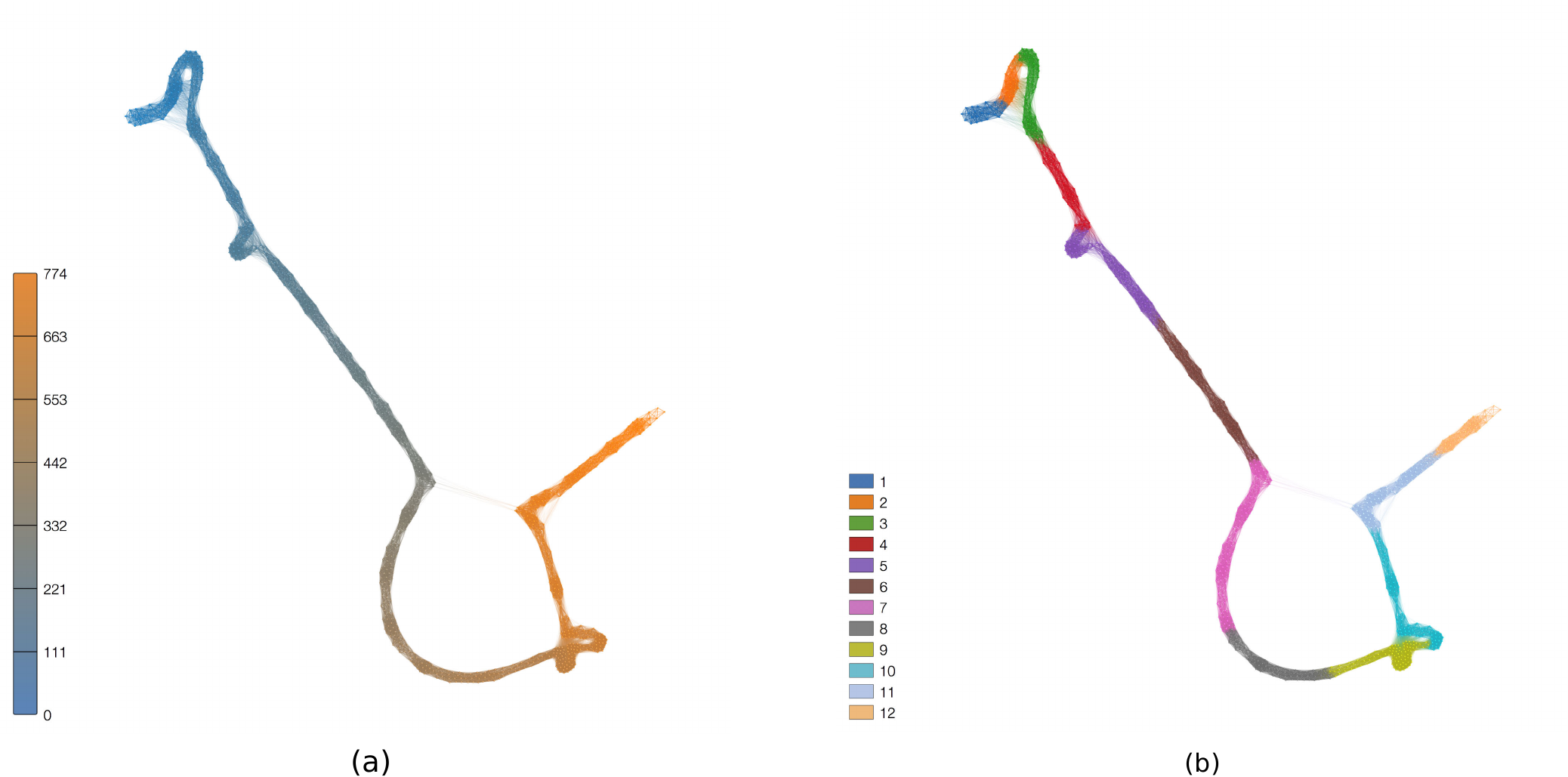}
  \caption{Visualization of the network $\mathcal{G}_1$, representing the book Alice's Adventures in Wonderland with a threshold $T_1=0.31$. Each node indicates a sequence of paragraphs. The order of the nodes according to the story is shown in (a). The first nodes of the story appear in blue, while the last nodes are represented in an orange color. In (b), the chapters of the nodes are represented with distinct colors.}
  ~\label{f:aliceNetworks}
  \end{center}
\end{figure*}

\begin{itemize}
 \item In chapter~1, we note that there is no strong connection among its paragraphs and those from other chapters, except for chapter~2 and 3, which is explained by the aforementioned overlap between subsequent paragraphs. The lack of long range connections among the nodes of the first chapter may happen because the main subject of this chapter is substantially different from almost every other in the book. In this chapter, the story unfolds in a more realistic scenario and it has no descriptions of the fantasy locations and creatures found in the rest of the book, except for the \emph{Rabbit};
 \item Chapters 2, 3, and the beginning of chapter 4 are connected among themselves. This may be a consequence of the fact that all these chapters describe the period of the story when Alice was very frightened of the world she has just jumped in. In addition, all these chapters mention when she cried and it formed a pool of tears;
 \item In chapter~5, there are strong connections between regions from the same chapter. This probably happens because there is a long conversation between Alice and the \emph{Caterpillar}, in which they discuss the many sizes she had during the previous chapters;
 \item The connection between chapters 7 and 11 can be related to the character \emph{The Hatter}, who is drinking tea in both chapters. Furthermore, he talked about specific kinds of food related to the tea party in both situations, e.g. bread and butter;
 \item There is a group of highly connected nodes in the end of chapter~9 and in the beginning of chapter~10. This probably happens because Alice met \emph{The Mock Turtle} in the last paragraphs of chapter 9 and their conversation ended only in chapter~10.
\end{itemize}

\begin{figure*}[ht!]
  \begin{center}
  \includegraphics[width=\linewidth]{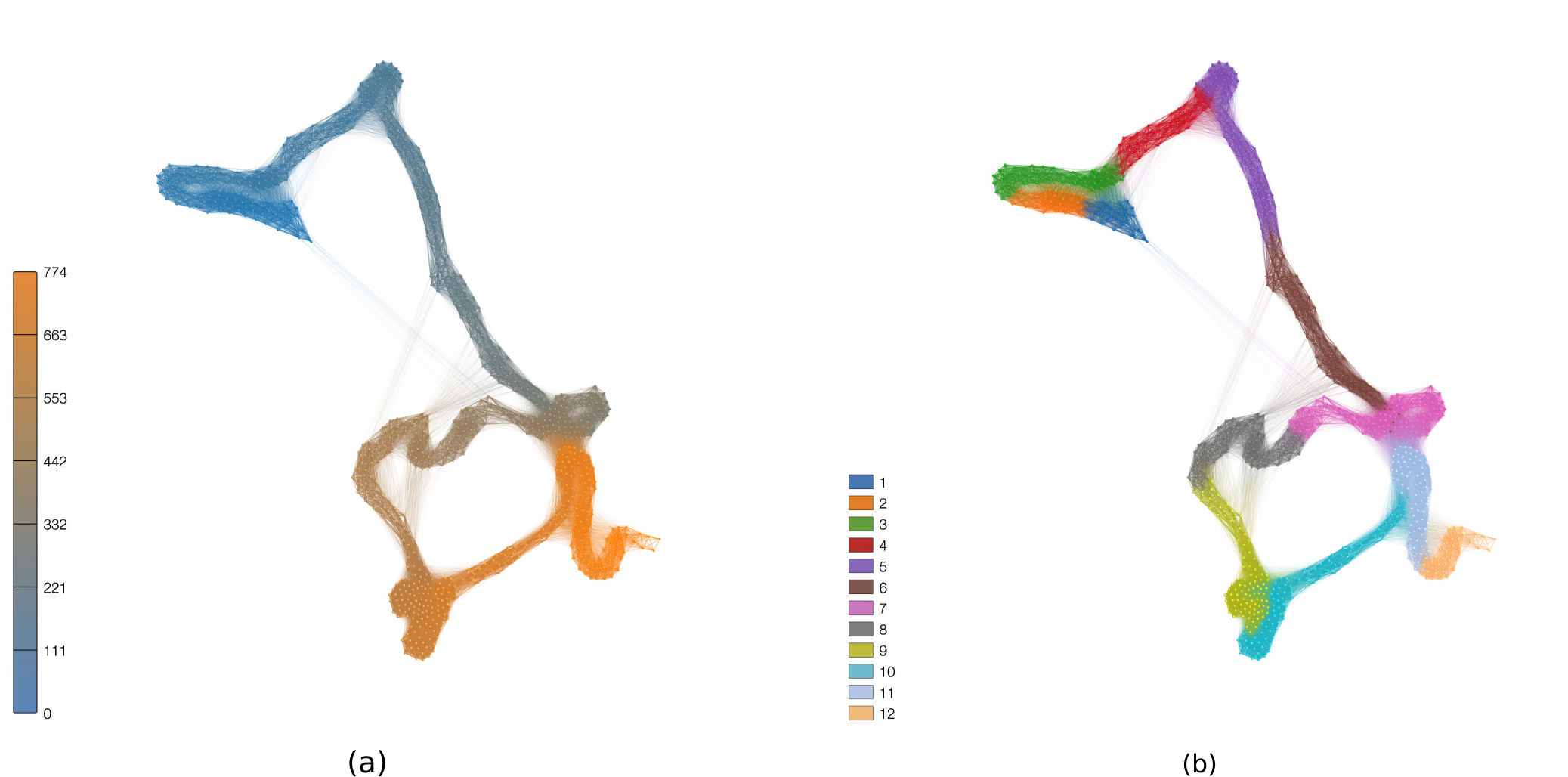}
  \caption{Visualization of the network $\mathcal{G}_2$, representing the book Alice's Adventures in Wonderland with a threshold $T_1=0.31$. The nodes indicate sets of 20 adjacent paragraphs. Item (a) shows the order of the nodes according to the story, where the first nodes of the story appear in blue and the last nodes in an orange. Item (b) represents the chapters, in which the nodes are represented with distinct colors.}
  ~\label{f:aliceNetworks2}
  \end{center}
\end{figure*}

Figure~\ref{f:aliceNetworks2} displays a visualization of the $\mathcal{G}_2$ network, which was constructed using a lower threshold value, $T_2=0.18$. By using two threshold choices, it has been possible to illustrate the potential of our method in describing the characteristics of the network in a multi-scale fashion. From Figure~\ref{f:aliceNetworks2}(a), we can observe the network still has a chain-like structure similar to that found in $\mathcal{G}_1$. However, this network presents more connections among nodes from different parts of the book. This is because the $\mathcal{G}_2$ network captures more fine-grained information about the relationships among the paragraphs. Comparing $\mathcal{G}_1$ and $\mathcal{G}_2$, we note that while chapter~1 is connected only with chapter~2 and 3 in $\mathcal{G}_1$, in $\mathcal{G}_2$ it  also connects with other parts of the book, in particular, with chapters 4 and 7. However, the analysis of fine-grained networks may present some disadvantages because these networks tend to incorporate more local characteristics. Moreover, they may include noise and relationships not driven by a strong contextual content.

\section{Discriminating real from shuffled texts}

To illustrate the ability of the proposed representation to grasp semantical information of texts by considering topological features, we evaluated the efficiency of the method in discriminating real texts from texts conveying no meaning, which are here represented by shuffled texts.  This is an important potential application of the proposed approach as a subsidy to fraud identification, such as inferring if texts in unknown languages are meaningful or not.

\begin{figure*}[ht!]
  \begin{center}
  \includegraphics[width=.97\linewidth]{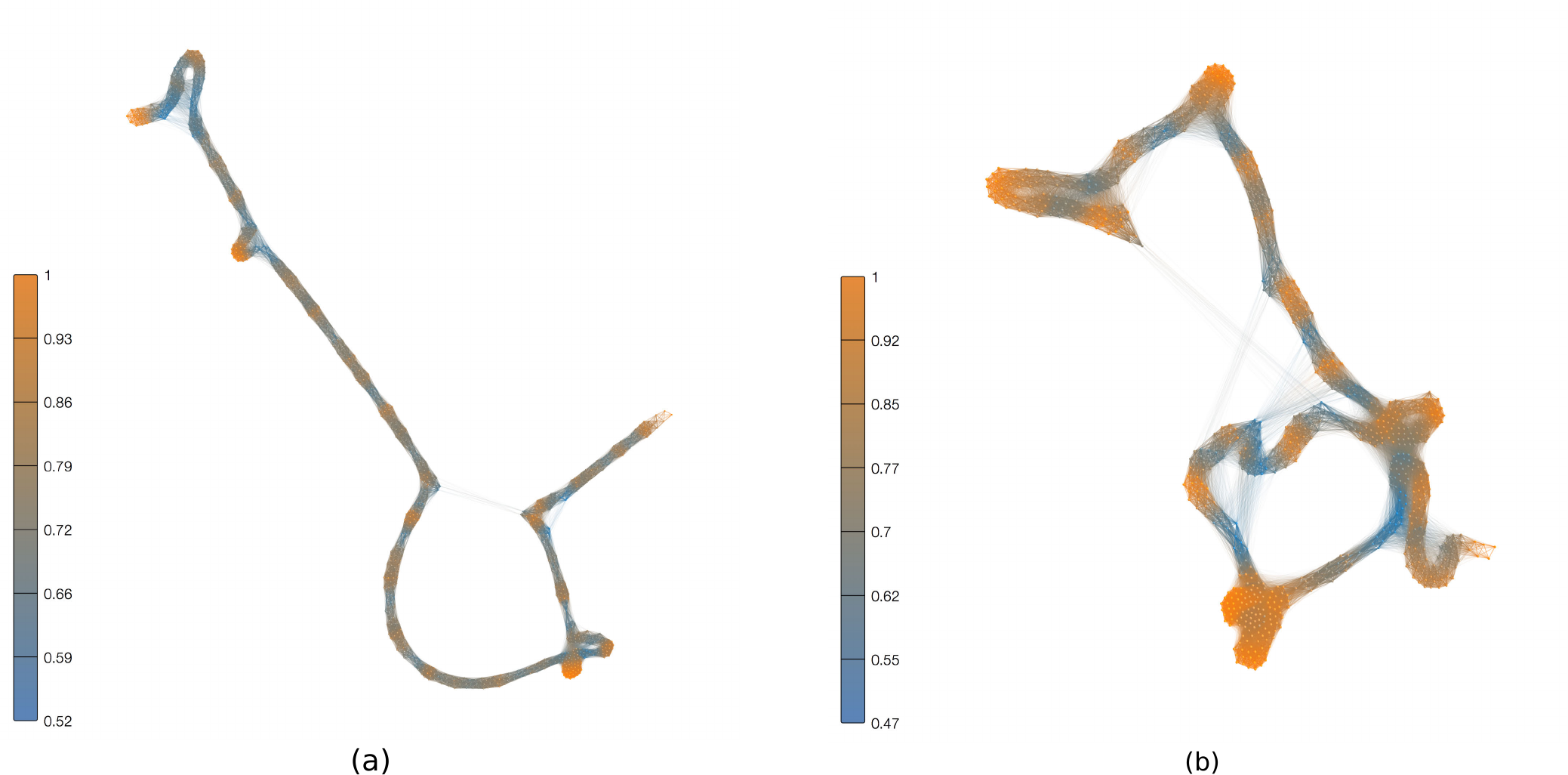}
  \caption{Visualization of the networks representing ``Alice's Adventures in Wonderland''. Item (a) represents the network $\mathcal{G}_1$ with a threshold $T_1=0.31$ and item (b) represents the network $\mathcal{G}_2$ with a threshold $T_2=0.18$. The node colors indicate the value of the clustering coefficient, in which nodes with the highest values are represented in orange. Note that there is a non-trivial pattern of clustering coefficient along the network nodes.}
~\label{f:aliceNetworksClustring}
\end{center}
\end{figure*}

\begin{figure}[ht!]
    \centering
    \subfigure[RT]{\includegraphics[width=0.5\textwidth]{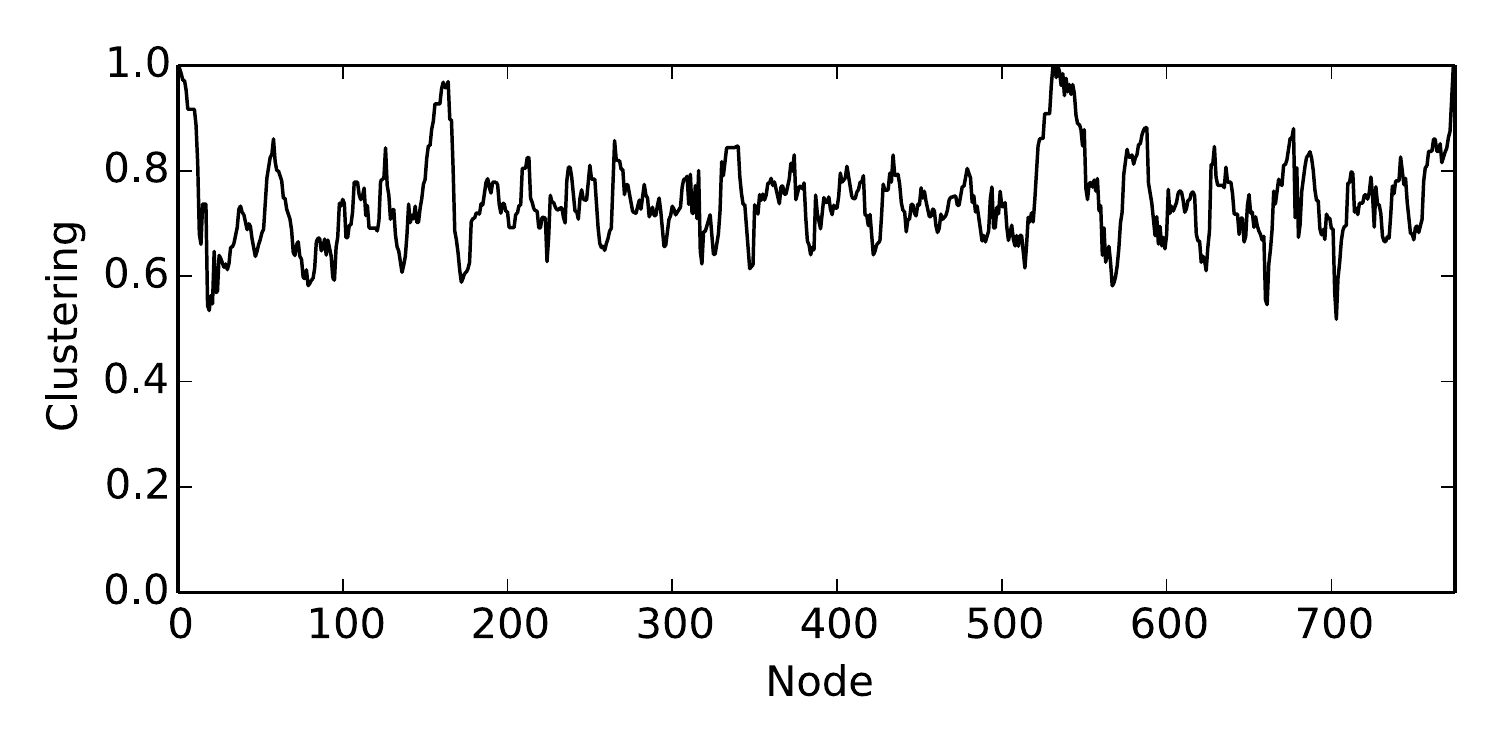}}
    \subfigure[SP]{\includegraphics[width=0.5\textwidth]{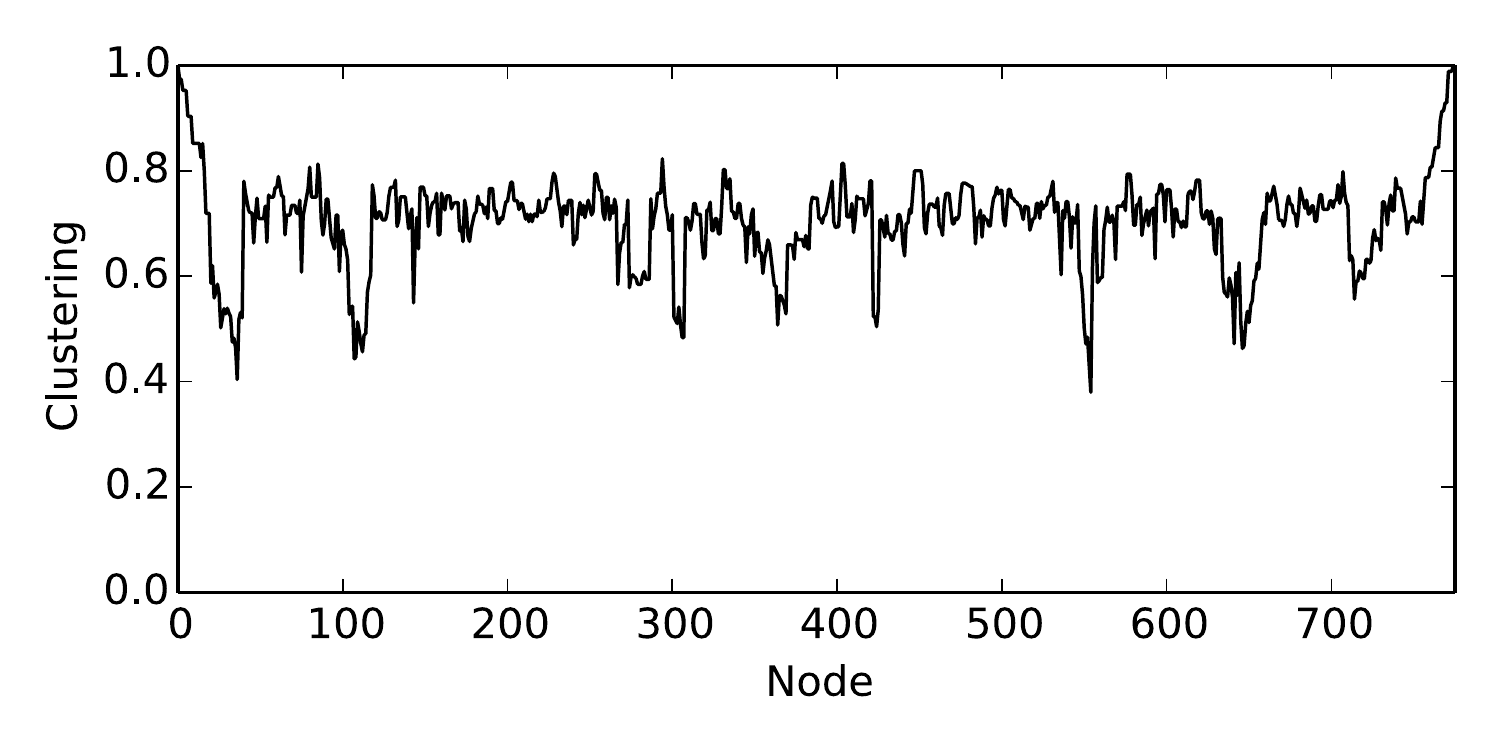}}
    \subfigure[SW]{\includegraphics[width=0.5\textwidth]{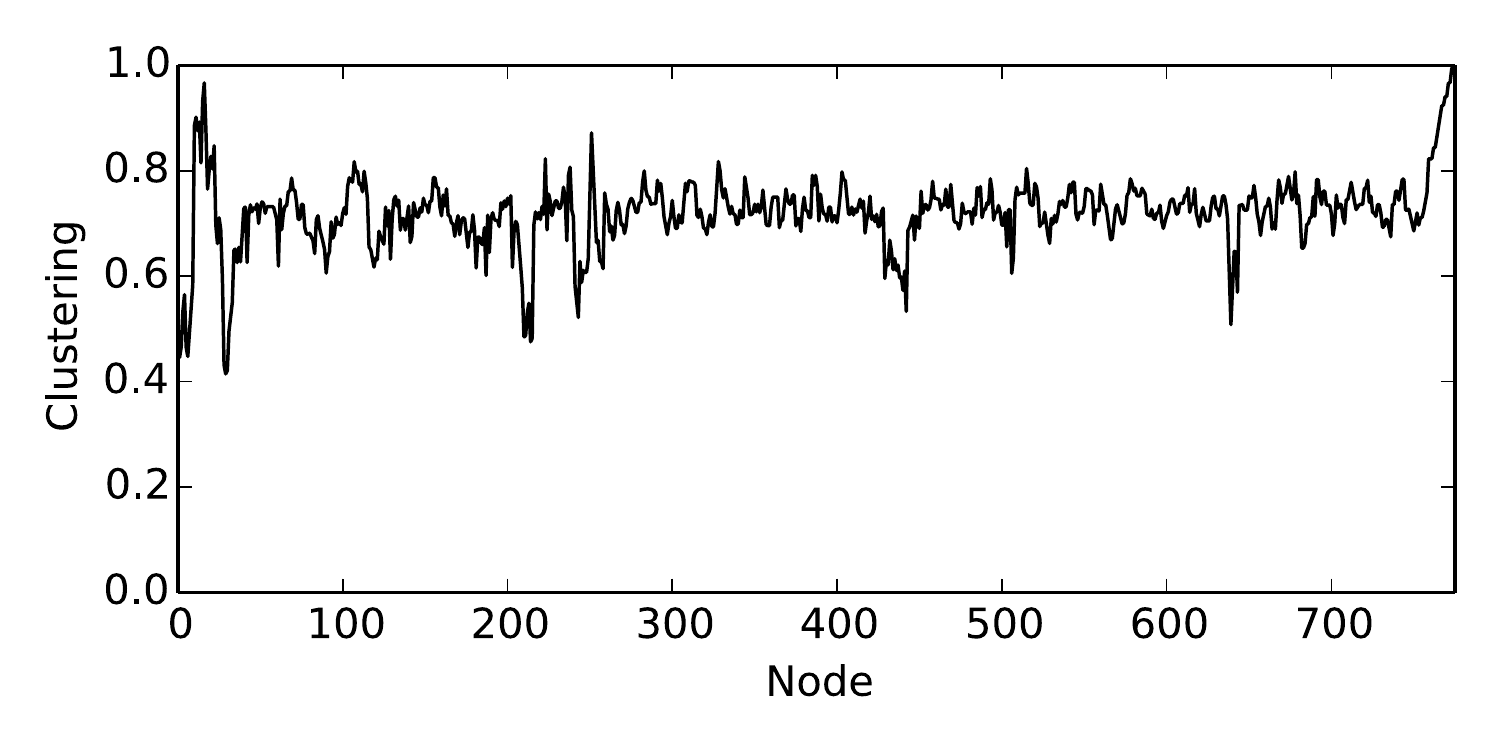}}
    \caption{Clustering coefficient for all network nodes of real and shuffled versions (RT, SW, and SP) created from  the book ``Alice's Adventures in Wonderland". The threshold $T_1=0.31$ was chosen to select the strongest semantical links.}\label{fig:clusteringXtime}
\end{figure}

In Figure \ref{f:aliceNetworksClustring}, we show the two networks obtained from the book `'Alice's Adventures in Wonderland'' and the respective values of clustering coefficient (for two thresholds) along the document. Note that an interesting pattern emerges in both cases. Regions encompassing many long-range connections are characterized by low values of clustering. In addition, there is a complex pattern of  intermittent appearences of low values of clustering coefficient in regions devoid of long-range connections. A similar behavior occurred with the matching index (result not shown). In Figure \ref{fig:clusteringXtime}, we show the behavior of the clustering coefficient along time for the real book and its two respective meaningless versions formed by shuffled paragraphs and words. It is clear from the figure that, in average, the clustering coefficient of all three versions fluctuates around $C\simeq0.78$. However, the patterns of fluctuations are markedly dissimilar. The largest variations arise for the real book, while both shuffled versions seem to display larger regions of weak fluctuations (see e.g. nodes from 180 to 280 in Figure \ref{fig:clusteringXtime}(b)). A similar pattern was obtained for the matching index measurement. Owing to the clear patterns in the fluctuations of local density discriminating real and meaningless texts, we applied measurements to quantify the mentioned fluctuations in order to check how much the proposed model depends on the text unfolding.

The fluctuations observed in Figure \ref{fig:clusteringXtime} were characterized with the coefficient of variation in a set of observations $X$, where $X$ here represents the ordered set of values of $C$ or $\mu$. The coefficient of variation ($c_v(X)$) is defined~\cite{das2008statistical} as:
\begin{equation}
	c_v(X) = {\sigma(X)}/{\langle X \rangle},
\end{equation}
where $\sigma(X)$ and $\langle X \rangle$ are the standard deviation and the average of $X$, respectively. For a choice of a window size, $\delta$, and for each possible subsequence of $X$, $\mathcal{X}^\delta_k = \{x_k, x_{k+1},\dots,x_{k+\delta-1}\}$, the coefficient of variation, $c_v(\mathcal{X}^\delta_k)$, is calculated. The set of $\delta$ values used in this paper was $\delta = \{3,5,7,10,15,20,25,30,35,40,50\}$. For each value of window size $\delta$, we summarize the values of fluctuations by averaging over all $c_v(\mathcal{X}^\delta_k$), i.e.:
\begin{equation}
 \mathcal{C}_v^\delta(X) = \frac{1}{N} \sum_{k=1}^{n-\delta+1} c_v(\mathcal{X}^\delta_k).
\end{equation}
Finally, each network was characterized by the set of features $\mathcal{F} = \{\mathcal{C}_v^{\delta=3},\mathcal{C}_v^{\delta=5},\mathcal{C}_v^{\delta=7}\ldots\}$, with $X$ being the values of clustering coefficient and matching index.

\begin{figure*}[ht!]
    \centering
    \subfigure[Mesoscopic networks.]{\includegraphics[width=0.49\textwidth]{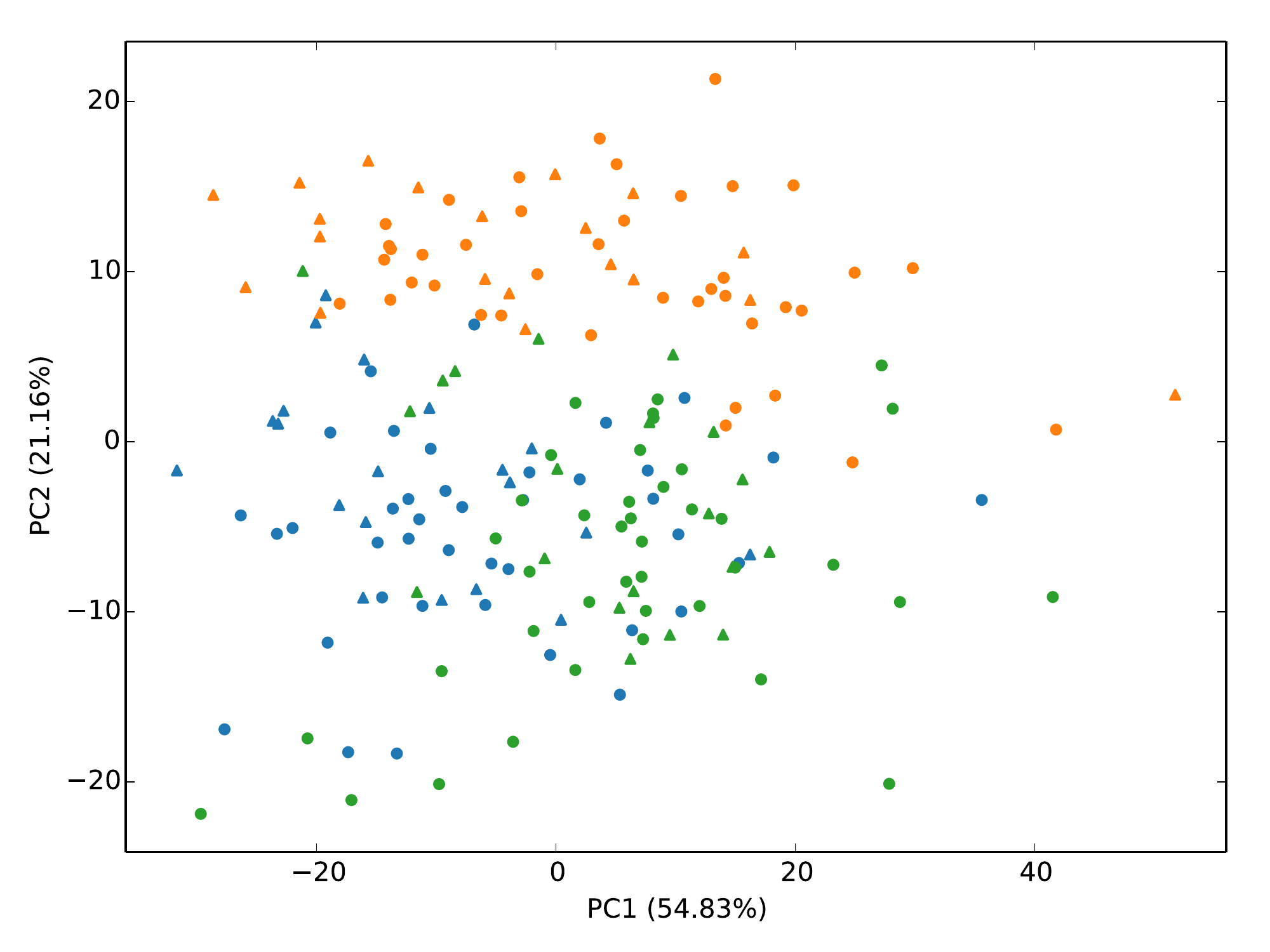}}
    \subfigure[Co-occurrence networks.]{\includegraphics[width=0.49\textwidth]{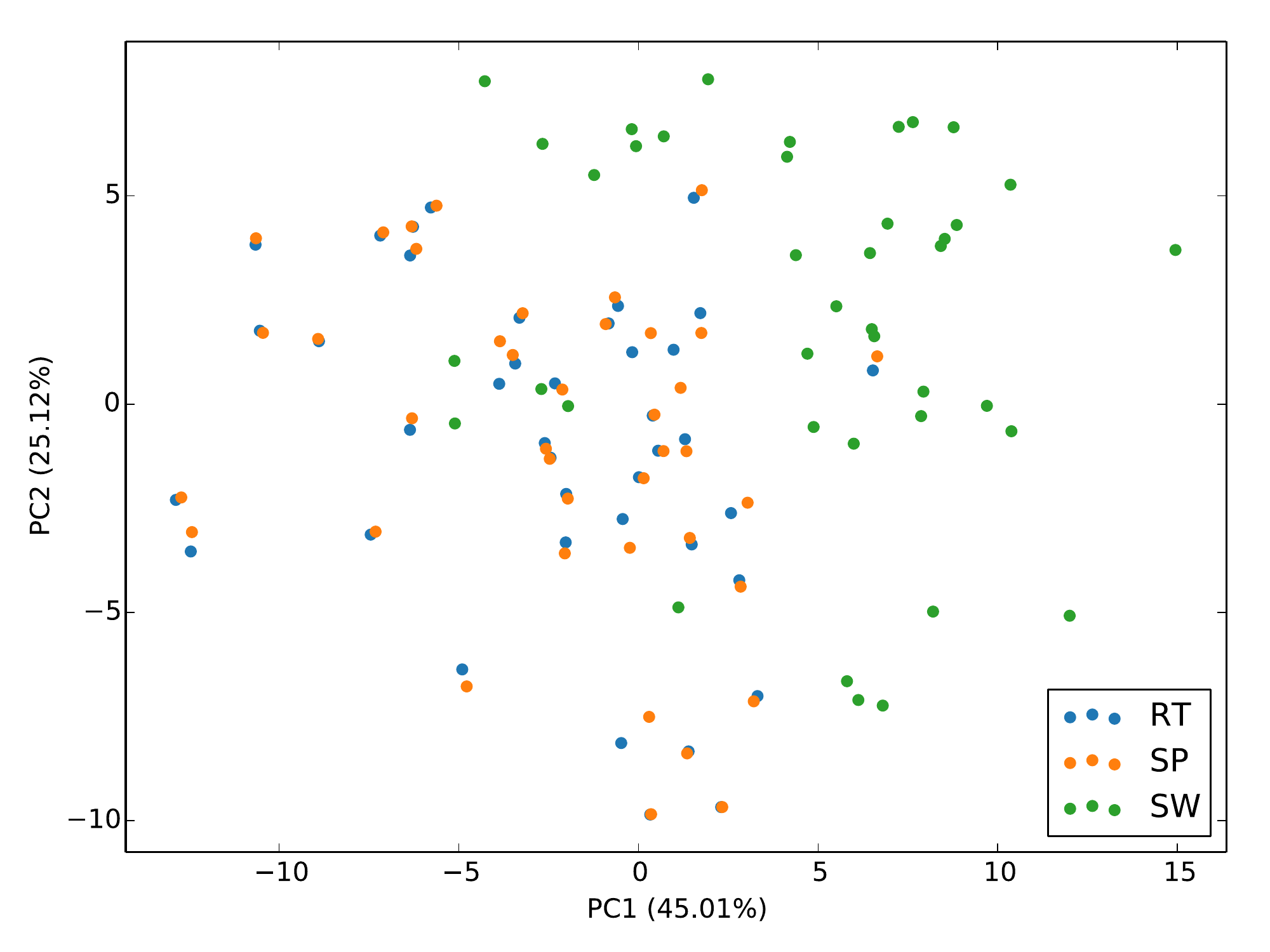}}
    \caption{PCA projections of the networks generated from Real Texts (RT), Shuffled Paragraphs (SP) texts, and Shuffled Words (SW) texts. The projections (a) and (b) represent the mesoscopic and the co-occurrence networks, respectively.}\label{pca}
\end{figure*}

To validate the potential of our mesoscopic model to extract the information from the document story, we considered the problem of discriminating real from meaningless (shuffled) texts using a dataset comprising several books (see details in Appendix \ref{ap.A}). We first visualized all three classes of texts in a bidimensional principal component analysis projection~\citep{jolliffe2002principal} (PCA). The results are shown in Figure \ref{pca}(a), in which the two first components account for approximately 76\% of the projection. Remarkably, the networks are usually placed close to others from the same class, while being well-separated from other classes. This latter effect is confirmed in terms of the average distance between classes shown in Table~\ref{tab:PCADistancesMesoscopic}. Our results are compared with those obtained with the traditional approach based on co-occurrence networks (see details in Appendix \ref{co-oc}). The PCA projection of these networks is shown in Figure~\ref{pca}(b). Although the sum of the two main PCA components accounts for 70\% of the projection, the group of networks from RT and SP are not distinguishable. This behavior was expected because co-occurrence networks were first devised to grasp linguistic/syntactical features. When language structure is kept and only the mesoscopic structure is changed (in SP texts), the co-occurrence approach is unable to discriminate real from meaningless texts. The poor discriminability observed is confirmed by the distances shown in  Table~\ref{tab:PCADistancesCoOccurence}.
\begin{table}[]
\centering
\caption{Average distance among networks from the same class. Note that, when using mesoscopic networks, it is possible to discriminate real texts from those generated by both shuffled words and paragraph. Conversely, if co-occurrence networks are used, real texts and texts formed by shuffled paragraphs cannot be discriminated.}
\label{tab:PCADistances}
\subtable[Mesoscopic network.]{
\begin{tabular}{|l|l|l|l|}
\hline
          & \multicolumn{1}{c}{RT}   & \multicolumn{1}{|c}{SW}      & \multicolumn{1}{|c|}{SP} \\ \hline
RT     & \multicolumn{1}{c}{0.00} & \multicolumn{1}{|c}{13.67}  & \multicolumn{1}{|c|}{10.98}  \\ \hline
SW    & \multicolumn{1}{c}{13.67} & \multicolumn{1}{|c}{0.00}  & \multicolumn{1}{|c|}{13.21}  \\ \hline
SP     & \multicolumn{1}{c}{10.98} & \multicolumn{1}{|c}{13.21}   & \multicolumn{1}{|c|}{0.00} \\ \hline
\end{tabular}
\label{tab:PCADistancesMesoscopic}
}
\subtable[Co-occurrence network.]{
\begin{tabular}{|l|l|l|l|}
\hline
          & \multicolumn{1}{c}{RT}   & \multicolumn{1}{|c}{SW}      & \multicolumn{1}{|c|}{SP} \\ \hline
RT     & \multicolumn{1}{c}{0.00} & \multicolumn{1}{|c}{7.18}  & \multicolumn{1}{|c|}{0.12}  \\ \hline
SW    & \multicolumn{1}{c}{7.18} & \multicolumn{1}{|c}{0.00}  & \multicolumn{1}{|c|}{7.08}  \\ \hline
SP     & \multicolumn{1}{c}{0.12} & \multicolumn{1}{|c}{7.08}   & \multicolumn{1}{|c|}{0.00} \\ \hline
\end{tabular}
\label{tab:PCADistancesCoOccurence}
}
\end{table}


The discriminability between real texts and the two classes of shuffled texts was also evaluated using an unsupervised approach based on the K-means algorithm~\citep{frank2009weka}. Here, we used the 6 principal components  as features as such choice yielded optimized results. Considering all documents of the datasets, only 8.9\% of instances were incorrectly clustered with the mesoscopic approach. Interestingly, the clustering generated by the algorithm yielded only 0.02\% of false negatives for the SP class. A feature relevance analysis revealed that the clustering coefficient outperforms the matching index for the clustering task, when the algorithm is applied using the measurements separately. When only the clustering and matching index are used, the percentage of incorrectly assigned instances are 11.7\% and 16.7\%, respectively.

The unsupervised approach was also used to compare the proposed methodology and traditional co-occurrence networks.   In this analysis, we used 10 principal components, as this amount of features yielded optimized results. The quality of clusters was estimated in terms of the accuracy the adjusted rand index (ARI)~\citep{hubert1985comparing}. The cluster quality indexes obtained in both types of networks are shown in Table~\ref{tab:performances}. Co-occurrence networks could not properly distinguish RT from SP classes, as expected from the analysis of Figure~\ref{pca}(b).  In this scenario, 72.5\% of SP texts were incorrectly classified as RT. This inability is also reflected in the ARI, which is much lower in co-occurrence networks.

\begin{table}[]
\centering
\caption{Comparison of the K-means clustering performance among different network approaches. Two different measurements were applied: Adjusted Rand Index (ARI) and Accuracy. In both measurements, 1 indicates that all instances are correctly classified and 0 indicates the opposite.}
\label{tab:performances}
\begin{tabular}{|l|l|l|}
\hline
                                  & \multicolumn{1}{c}{ARI}   & \multicolumn{1}{|c|}{Accuracy} \\ \hline
Mesoscopic (Clustering)     & \multicolumn{1}{c}{0.679} & \multicolumn{1}{|c|}{0.883}    \\ \hline
Mesoscopic (Matching Index) & \multicolumn{1}{c}{0.576} & \multicolumn{1}{|c|}{0.833}    \\ \hline
Mesoscopic (all features)                      & \multicolumn{1}{c}{0.749} & \multicolumn{1}{|c|}{0.911}    \\ \hline
Co-occurrence                     & \multicolumn{1}{c}{0.268} & \multicolumn{1}{|c|}{0.575}    \\ \hline
\end{tabular}
\end{table}

\begin{figure*}[ht!]
    \centering
    \subfigure[RT]{\includegraphics[width=0.3\textwidth]{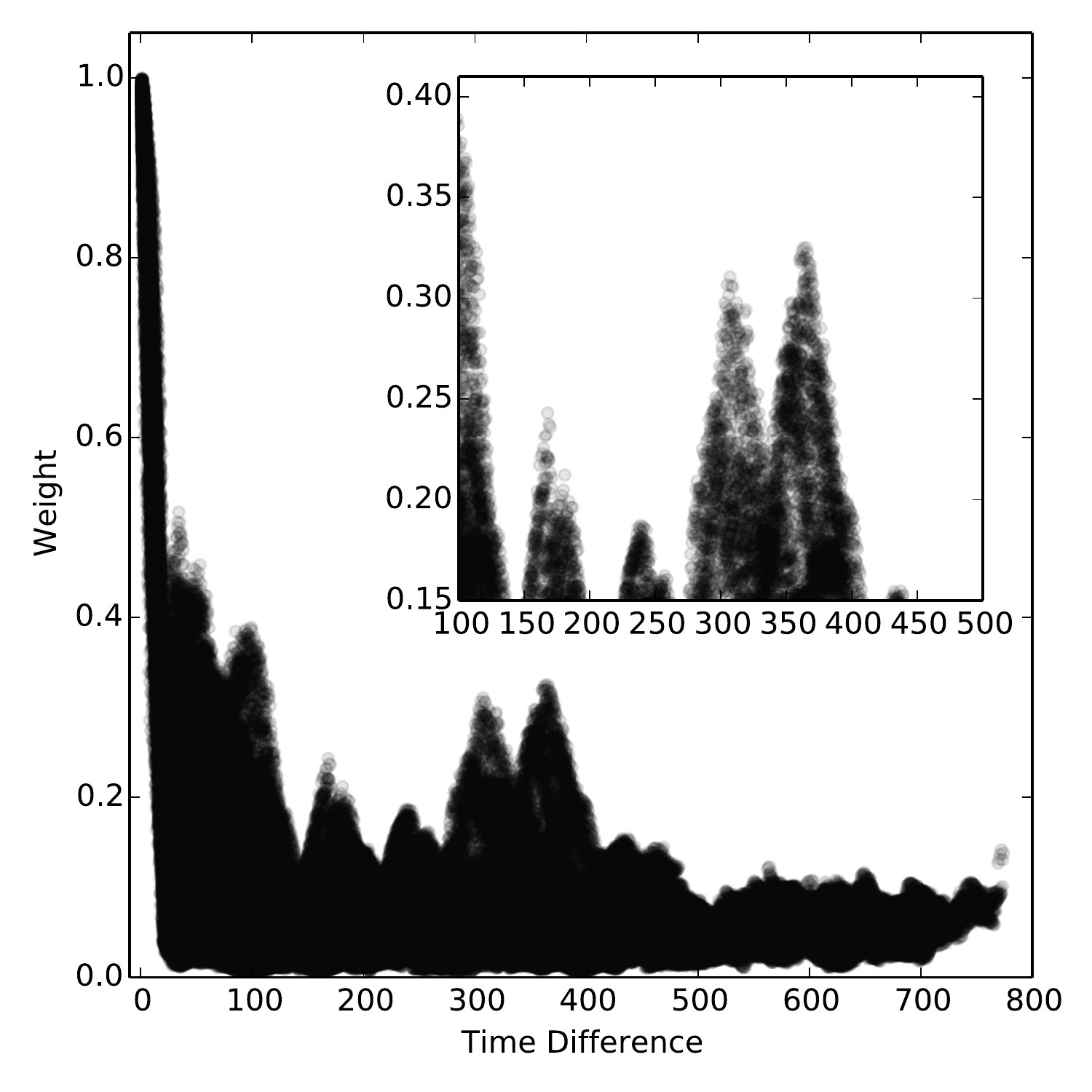}}
    \subfigure[SP]{\includegraphics[width=0.3\textwidth]{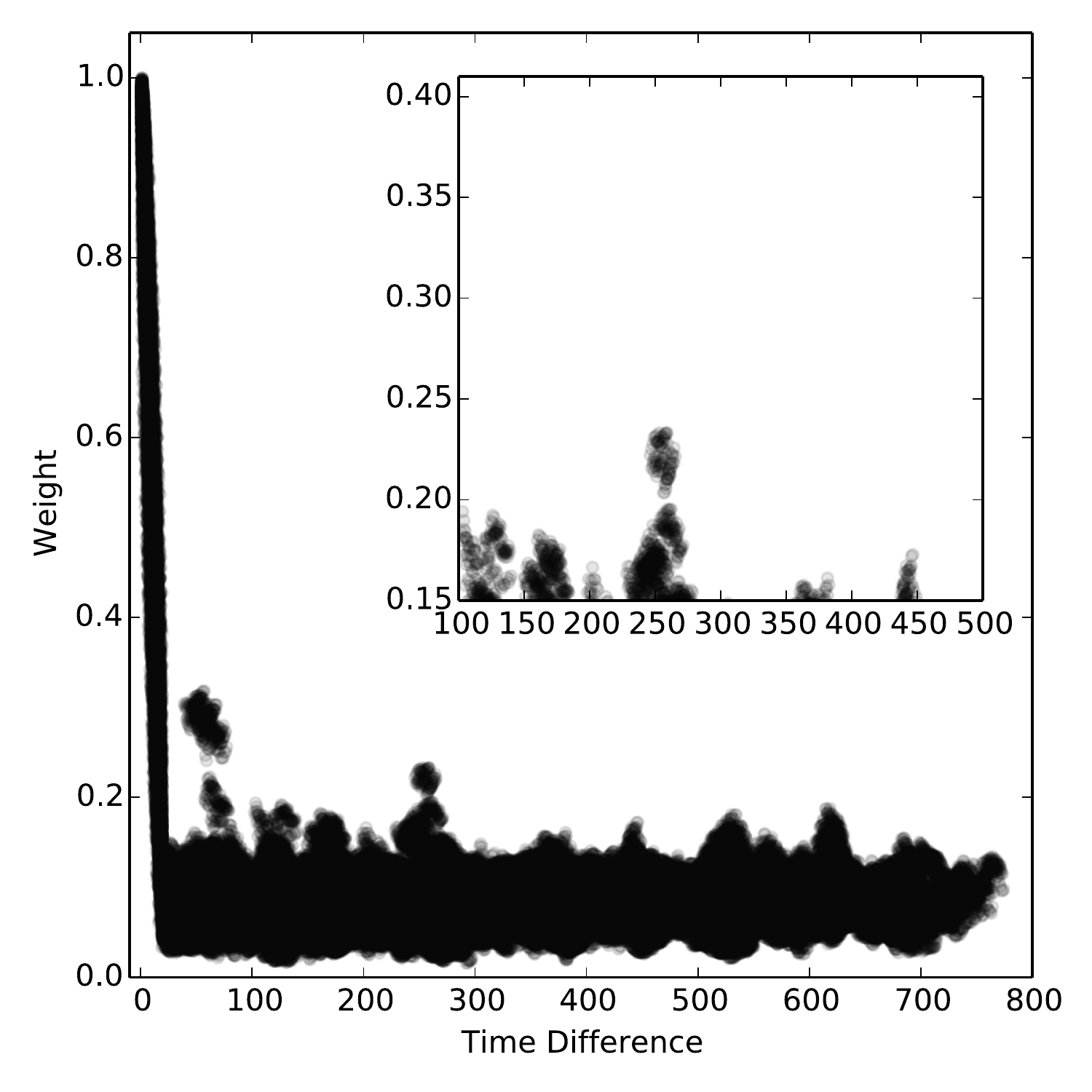}}
    \subfigure[SW]{\includegraphics[width=0.3\textwidth]{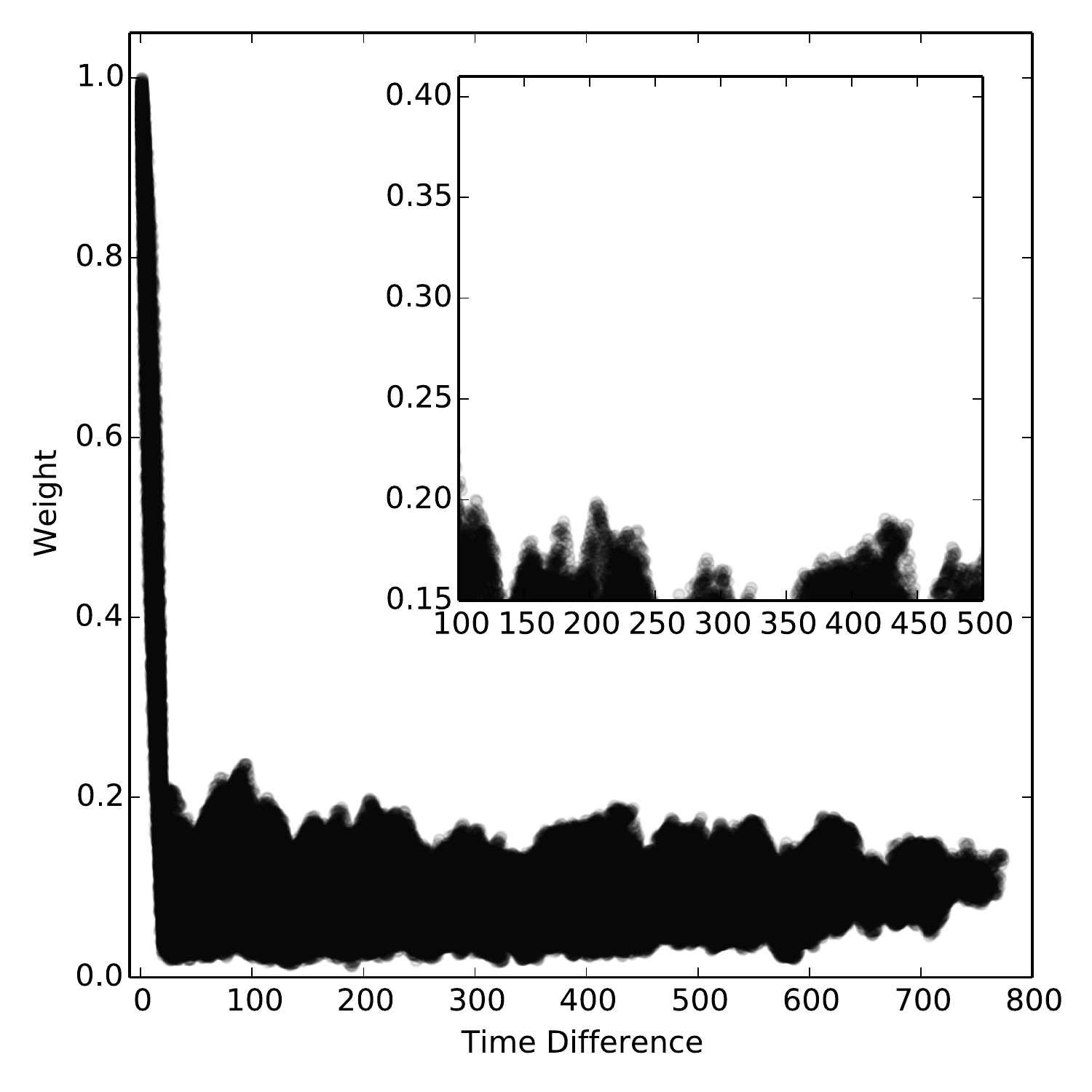}}
    \caption{Time difference between linked nodes vs. edges weights. This measurement was computed for all network edges and the inset represents a region of long-range links, i.e. links with time difference larger than 100. Comparing the different classes of texts, it is evident that strong long-range connections are more likely to appear in real networks.} \label{fig:weightXtime}
\end{figure*}

A particular feature of the mesoscopic model is the existence of long-range connections. More specifically, a long-range connection is a link that connects two nodes that are far apart in the document. This type of link usually appears when a subject/context previously mentioned in the book is revisited in the story. It has been conjectured that such links, a consequence of the long-range correlation effect~\citep{EBELING1995233}, are essential for mapping a multidimensional conceptual space into a smaller dimensional space~\citep{Alvarez-Lacalle23052006}.   To quantify the presence of long-range links, we show (Figure~\ref{fig:weightXtime}) scatterplots of edges weights versus the time difference between linked nodes, where time corresponds to the natural reading order. In all three classes of texts, as imposed by the  construction rules of mesoscopic networks, many edges are established between successive nodes.
Long-range connections were also observed in the three classes of texts (i.e. RT, SW, and SP). However, most of such long connections are very weak. As depicted in the inset of Figure \ref{fig:weightXtime}, real texts tend to present stronger long range connections than shuffled texts, especially in the time frame of 300 to 400 paragraphs.

\section{Conclusion}

In order to grasp semantical, mesoscopic properties of texts modeled as networks, we proposed an approach that considers the semantical similarity between textual segments. Differently from previous representations, we modeled sequences of adjacent paragraphs as nodes, whose links are established by content similarity. By doing so, we could capture two important features present in written texts: long-range correlations and the temporal unfolding of documents. In addition, the proposed approach for text representation also allowed multi-scale representation of documents.  Specifically, two parameters control the scale: (i) $\Delta$: the number of consecutive paragraphs in each window, and (ii) $T$: the threshold used to prune connections among nodes with low contextual similarity.

As a case study, we tested our approach in ``Alice's Adventures in Wonderland'', by employing network visualization techniques on the generated mesoscopic network. Many insights could be drawn from the visualization by tracing a parallel between its underlying structure and the story. In particular, we investigated the correspondence between the content of each chapter and the underlying network structure arising from the proposed model.
Our model uncovered many relationships among different contexts sharing the same topics, such as similar characters or places throughout the story. For example, the high contextual similarity found between chapters~7 and 11 can be explained by the fact that both chapters share a recurrent subject revolving around the character \emph{The Hatter} and the tea party thematic. Note that similar textual inferences could not be drawn from models solely based on local features, as it is the case of traditional word-adjacency or syntactical networks, as they emphasize mostly stylistic textual subtleties.

The effectiveness of our model was also evaluated with respect to the task of discriminating real from shuffled texts. The shuffled versions, particularly, were created by mixing either words or paragraphs of real texts. We have found that, if we consider only two simple local density measurements, it is possible to separate all three classes of texts with high accuracy. The traditional co-occurrence turned out to grasp only local subtleties, as the model was not able to discriminate real texts from those generated by shuffling paragraphs.  This happens because, when paragraphs are shuffled, only a few edges -- those at the paragraph boundaries -- are modified. These results confirm the suitability of the proposed model in capturing larger contexts in a mesoscopic fashion. A further analysis of the model also revealed that real texts are characterized by stronger long-range links, a feature that could be explored in tests of informativeness of written documents~\citep{10.1371/journal.pone.0067310}.

The proposed network representation paves the way for developing new techniques that could be applied to automatically analyze the mesoscopic structure of documents. These techniques could improve traditional approaches used to tackle typical text mining problems under a new perspective. This capability should be further explored in future works, for instance, by measuring the efficiency of our model in text classification, summarization and similar applications in which an accurate semantic analysis plays a prominent role in the characterization of written texts.

\section*{Acknowledgements}

The authors acknowledge financial support from Capes-Brazil,  S\~ao Paulo Research Foundation (FAPESP) (grant no. 2016/19069-9,
2015/08003-4, 2015/05676-8, 2014/20830-0 and 2011/50761-2), CNPq-Brazil (grant no. 307333/2013-2) and NAP-PRP-USP.

\newpage

\ \\

\newpage
 
\onecolumngrid
 
\appendix

\section{Dataset} \label{ap.A}

All the texts used in our dataset were extracted from the open access Project Gutemberg dataset~\footnote{Project Gutemberg - https://www.gutenberg.org/}. We divided the dataset into two major groups, according to the original language: (i)~English and (ii)~Other languages. The books, sorted by language and author, are listed below:

English:
 \begin{itemize}
  \item \textbf{Arthur Conan Doyle}: \emph{The Adventures of Sherlock Holmes}; \emph{The Tragedy of the Korosko}; \emph{The Valley of Fear}; \emph{Through the Magic Door} and \emph{Uncle Bernac - A Memory of the Empire};
  \item \textbf{Bram Stoker}: \emph{Dracula's Guest}; \emph{The Lair of the White Worm}; \emph{The Jewel Of Seven Stars}; \emph{The Man} and \emph{The Mystery of the sea};
  \item \textbf{Charles Dickens}: \emph{A Tale of Two Cities}; \emph{American Notes}; \emph{Barnaby Rudge: A Tale of the Riots of Eighty}; \emph{Great Expectations} and \emph{Hard Times};
  \item \textbf{Edgar Allan Poe}: \emph{The Works of Edgar Allan Poe (Volume 1 - 5)};
  \item \textbf{Hector H. Munro (Saki)}: \emph{Beasts and Super-Beasts}; \emph{The Chronicles of Clovis}; \emph{The Toys of Peace}; \emph{When William Came} and \emph{The Unbearable Bassington};
  \item \textbf{P. G. Wodehouse}: \emph{The Girl on the Boat}; \emph{My Man Jeeves}; \emph{Something New}; \emph{The Adventures of Sally} and \emph{The Clicking of Cuthbert}
  \item \textbf{Thomas Hardy}: \emph{A Pair of Blue Eyes}; \emph{Far from the Madding Crowd}; \emph{Jude the Obscure}; \emph{The Mayor of Casterbridge} and \emph{The Hand of Ethelberta};
  \item \textbf{William M. Thackeray}: \emph{Barry Lyndon}; \emph{The Book of Snobs}; \emph{The History of Pendennis}; \emph{The Virginians} and \emph{Vanity Fair}
 \end{itemize}
Other languages:
 \begin{enumerate}
  \item French:
	\begin{itemize}
	  \item \textbf{Gustave Aimard}: \emph{Le fils du Soleil};
	  \item \textbf{Jules Verne}: \emph{Face au Drapeau};
      \item \textbf{Louis Am\'ed\'ee Achard}: \emph{Pierre de Villergl\'e};
	  \item \textbf{Louis Reybaud}: \emph{Les Idoles d'argile};
	  \item \textbf{Victor Hugo}: \emph{Han d'Islande}.
	\end{itemize}
  \item German:
	\begin{itemize}
	  \item \textbf{Goethe}: \emph{Die Wahlverwandtschaften};
	  \item \textbf{Jakob Wassermann}: \emph{Der Moloch};
	  \item \textbf{Robert Walser}: \emph{Geschwister Tanner};
	  \item \textbf{Thomas Mann}: \emph{K\"onigliche Hoheit};
	  \item \textbf{Wilhelm Hauff}: \emph{Lichtenstein}.
	\end{itemize}
  \item Italian
	\begin{itemize}
	  \item \textbf{Alberto Boccardi}: \emph{Il Peccato di Loreta};
	  \item \textbf{Anton Giulio Barrili}: \emph{La Montanara};
	  \item \textbf{Enrico Castelnuovo}: \emph{Alla Finestra};
	  \item \textbf{Guido da Verona}: \emph{Sciogli la treccia, Maria Maddalena};
	  \item \textbf{Virginia Mulazzi}: \emph{La Pergamena Distrutta}.
	\end{itemize}
  \item Portuguese:
	\begin{itemize}
	  \item \textbf{Camilo Castelo Branco}: \emph{Amor de Perdi\c{c}\~ao};
	  \item \textbf{E\c{c}a de Queir\'os}: \emph{A cidade e as Serras};
	  \item \textbf{Faustino da Fonseca}: \emph{Os Bravos do Mindello};
      \item \textbf{Jaime de Magalh\~aes Lima}: \emph{Transviado};
	  \item \textbf{J\'ulio Dinis}: \emph{ Uma Fam\'ilia Inglesa}.
	\end{itemize}
 \end{enumerate}

\section{Characterization of co-occurrence networks} \label{co-oc}

Tipically, co-occurrence (or word adjacency) networks are formed by mapping each concept into a distinct node of the network. The edges are established by adjacency relationships, i.e. if two words are adjacent in the text, they are connected in the network. Such networks have been extensively explored in the context of text analysis and pattern recognition~\citep{masucci2006network}.
In the present work, we compare the properties of the mesoscopic and co-occurrence models.

We compare the mesoscopic results with a set of centrality measurements of co-occurrence networks used in the ref.~\citep{de2016using}, which are: accessibility~\citep{travenccolo2008accessibility}, betweenness centrality~\citep{freeman1977betweenness}, closeness centrality, clustering coefficient, degree, eccentricity~\citep{estrada2012structure}, eigenvector centrality~\citep{bonacich1987power}, generalized accessibility~\citep{arruda2014role}, modularity~\citep{newman2004finding} (computed from fast greedy algorithm~\citep{clauset2004finding}), neighborhood connectivity, number of nodes, PageRank~\citep{langville2011google}, and, symmetry~\citep{de2016using}. Apart from modularity, we compute the following quantities for each measurement: maximum value ($\max(X)$), median ($\tilde X$), minimum value ($\min(X)$), and standard deviation $\sigma(X)$. To create the co-occurrence networks, we trimmed the texts to the same number of words because many of the above complex network measurements are influenced by the number of nodes. Because the number of network nodes varies in mesoscopic networks, we did not use the same set of measurements as for the co-occurrence networks. Furthermore, in the co-occurence network analysis, we only used texts written in English because this kind of representation catches information regarding the syntax, which is different for each language.

\newpage

\bibliographystyle{apalike}


%
%
%
%
%

\end{document}